\documentclass[conference]{IEEEtran}
\IEEEoverridecommandlockouts

\usepackage{multirow}
\usepackage{subfigure}
\usepackage{amsmath,amssymb,amsfonts}
\usepackage{algorithmic}
\usepackage{graphicx}
\usepackage{textcomp}
\usepackage{xcolor}
\usepackage{booktabs}
\usepackage[round, sort, numbers]{natbib}
\usepackage{url}

\def\BibTeX{{\rm B\kern-.05em{\sc i\kern-.025em b}\kern-.08em
    T\kern-.1667em\lower.7ex\hbox{E}\kern-.125emX}}
\begin{document}

\title{Personality Detection and Analysis using Twitter Data}

\author{

\IEEEauthorblockN{Abhilash Datta}
\IEEEauthorblockA{\textit{Department of Computer Science } \\
\textit{Indian Institute of Technology}\\
Kharagpur, India \\
abhilashdatta8224@gmail.com}
\and
\IEEEauthorblockN{Souvic Chakraborty}
\IEEEauthorblockA{\textit{Department of Computer Science } \\
\textit{Indian Institute of Technology}\\
Kharagpur, India \\
chakra.souvic@gmail.com}
\and
\IEEEauthorblockN{Animesh Mukherjee}
\IEEEauthorblockA{\textit{Department of Computer Science } \\
\textit{Indian Institute of Technology}\\
 Kharagpur, India\\
animeshm@gmail.com}}

\maketitle

\begin{abstract}
Personality types are important in various fields as they hold relevant information about the characteristics of a human being in an explainable format. They are often good predictors of a person's behaviors in a particular environment and have applications ranging from candidate selection to marketing and mental health. Recently automatic detection of personality traits from texts has gained significant attention in computational linguistics. Most personality detection and analysis methods have focused on small datasets making their experimental observations often limited. To bridge this gap, we focus on collecting and releasing the largest automatically curated dataset for the research community which has 152 million tweets and 56 thousand data points for the Myers-Briggs personality type (MBTI) prediction task. We perform a series of extensive qualitative and quantitative studies on our dataset to analyze the data patterns in a better way and infer conclusions. We show how our intriguing analysis results often follow natural intuition. We also perform a series of ablation studies to show how the baselines perform for our dataset.
\end{abstract}


\section{Introduction}
The personality of an individual refers to the specific collection of psychological constructs which dictates the visible differences in different human beings in terms of behavior and reaction in particular environments and also dictates the thought process which leads to these different behavioral outcomes (as defined in \citet{roberts2008personality}). Many researchers have recently tried automatic personality detection with little success primarily because the task is inherently difficult, requiring a thorough understanding of sentence constructs, sentiment toward targets, and its connection to behavioral outcomes. Sentiment analysis alone can be a very challenging task due to abundance of aspects and sparsity of labelled data\cite{souvic_aspect,souvic_hate}. Moreover, most research has been carried out on small datasets. Since the expression of a specific personality can have a wide range, small datasets often are unable to capture this variety and thus fail to provide the model with a sufficient inductive bias to learn.

Furthermore, the models used till now lack task-specific design which is essential to solving a complex problem. Attempts for personality modeling ranged from traditional methods like questionnaires to NLP-based approaches. The two widely-used personality models are the Big five personality traits (OCEAN model), coming from Sir Francis Galton's line of work (as described in \citet{goldberg1993structure,rothe2017scientific,rushton1990sir}) based on linguistically predictive personality types having 5 personality dimensions and the Myers-Briggs Type Indicator (MBTI) personality modeling, based on Carl Jung's theory, containing four personality dimensions as proposed in \citet{jung1971personality}. While there has been considerable work with the first kind of personality types being invented to be used by linguists, works on MBTI personality types are lacking. We hope to bridge this gap by introducing the largest automatically collected dataset for MBTI personality types. Our contributions in this paper are as follows.
\begin{enumerate}
    \item We introduce the largest dataset for personality detection with MBTI personality types. We perform all our analyses and automatic classification using the functional personality groups. However, we opensource this dataset in the original form with nuanced attributes containing all the individual 16 personalities for the community to fuel further research and exploration.
    \item We perform several quantitative and qualitative studies to analyze the dataset. We introduce novel features like \textbf{hashtags, URLs, and Mentions embeddings}, and show how they correlate with an individual's personality. We analyze personality types in several derivative dimensions like professions, readability, and empath features.
    \item We test several machine learning models on the task of predicting MBTI personality types from Twitter profile data. We fine-tune different models taking individual inputs to make better embeddings and use these embeddings to train another model finally enhancing the prediction accuracy. The best accuracy is achieved by a simple random forest classifier over fastText embeddings. 

    \item We perform a series of ablation studies to understand which features are important for the task. We show that the \textbf{hashtags} used by the users, their \textbf{empath} features, and their \textbf{tweets} are the most important features. We also show the impact of data quality and the number of tweets on the model's prediction performance.
\end{enumerate}

\section{Related Works}
Personality information can be valuable for a number of applications. Numerous research papers related to predicting personality traits among social media networks have recently surged interest in the research community\cite{indira2021personality,vstajner2020survey, mehta2020recent}. Previous research on the prediction of personality uses Twitter, Instagram, and Facebook data include some feature-based techniques such as LIWC\cite{holtgraves2011text}, SPLICE (structured programme for linguistic cue extraction)\cite{tadesse2018personality}, SNA (social network analysis) \cite{clifton2017introduction}, as well as time-based features\cite{wang2018sensing}. \citet{mitchell2015quantifying} studied self-identified schizophrenia patients on Twitter and found that linguistic signals may aid in identifying and getting help to people suffering from it.
\citet{luyckx2008personae} presented a corpus for computational stylometry, including authorship attribution and MBTIs for Dutch. The corpus consists of 145 students (BA level) essays. They controlled for the topic by asking participants to
write about a documentary on artificial life. In a
follow-up study \cite{verhoeven2014clips}, they extended the corpus to include reviews and both Big Five and MBTI information. Instead, we focus on English and social media, a
more spontaneous sample of language use.
Even when using social media, most prior
work on personality detection can be considered small-scale. The 2014 Workshop on Computational Personality Recognition hosted a shared task of personality detection on 442 YouTube video logs\cite{celli2014workshop}. \citet{celli2013workshop}
also examined Facebook messages of 250 users
for personality. \\
In contrast, \textbf{our study uses 152M
tweets from 56K different users}.
The only two prior large-scale open-vocabulary
works on social media study Facebook messages (\cite{park2015automatic,preoctiuc2015role}). To date, these studies
represent the largest one connecting language and personality. They collected
personality types and messages from 75,000 Facebook users through a Facebook app. They found striking variations in
language use with personality, gender, and age.
On the other hand, we collect our data using the Twitter API and in an automated way, retrieving every possible detail. We also generate our own set of features from the tweets for better classification. Our approach is simpler, requires no tailored app,
and can be used to collect large amounts of auto-annotated data
quickly. 

\section{Dataset Creation}
The most popular dataset on MBTI personality detection from text has only 1,500 data points with 1.2M tweets\cite{plank-hovy-2015-personality}. We attempt to create a new dataset containing tweets, user descriptions (bio), profile metadata (follower count, media count, listed count, etc.), and finally the MBTI personality type for each user through automatic means. 

\subsection{Data collection procedure}
We collect data from people who have publicly shared their personality test results (from \url{www.16personalities.com}) on Twitter. Our data collection strategy is as follows.

\begin{enumerate}
    \item \textbf{User mapping from profile link}: Complete profile links for the website follow a specific pattern of \url{https://www.16personalities.com/profiles/id} where \textit{id} refers to the id of the profile of that person. We use \textbf{Twitter API} to search for this pattern and obtain all the users who have shared their profile links. Then we use \textbf{selenium} to collect the personality type results calculated from the link.

    \item \textbf{User mapping from MBTI links}:
Some people, instead of sharing their test results directly, share the links to their respective personality types. To capture this, we search for links with pattern \url{https://www.16personalities.com/ptype-personality} where \textit{ptype} is the four-dimensional personality type. We collect this data using the Twitter API and filter out the cases where the same person has ever shared different personality type links.

\item \textbf{Collection of tweets}:
Twitter makes the last 3200 tweets for each user available in its web API. We use \textbf{snscrape} to collect the same for each user obtained in steps (1) and (2) and save them in text files in order to use them as input data.

\item \textbf{Collection of descriptions and metadata}:
From the profiles retrieved from step (1) and step (2), we use Twitter's Python API \textbf{Tweepy} to map the profile usernames to their Twitter ids. Then we retrieve their user objects containing description (bio) and other profile metadata like follower count, friend count, media count, etc. using \textbf{snscrape}. 

\end{enumerate}


\begin{table}
\centering
\renewcommand{\arraystretch}{1.0}
\begin{tabular}{|c|c|c|c|}
\hline
\textbf{Personality class} & \textbf{Personality type} & \textbf{\#Users} &\textbf{\#Tweets}\\
\hline
\multirow{5}{*}{Analyst} 
& \texttt{intp}	&2,462	&6,454,286\\
 & \texttt{intj}	&2,570	&6,685,845\\
& \texttt{entj}	&1,524	&4,102,830\\
& \texttt{entp}	&1,214	&4,655,466\\
\cline{2-4}
& \textbf{Subtotal} &\textbf{ 7,770} & \textbf{21,898,427}\\
\hline

\multirow{5}{*}{Diplomats} 
& \texttt{enfj}	&4,413	&11,343,062 \\
& \texttt{infj}	&6,728	&25,003,695\\
& \texttt{infp}	&9,125	&19,655,881\\
 & \texttt{enfp} 	&13,369 &34,696,360\\
 \cline{2-4}
 & \textbf{Subtotal} & \textbf{33,635} & \textbf{90,698,998}\\

\hline

\multirow{5}{*}{Sentinels} 
& \texttt{estj}	&547 &1,437,151\\
& \texttt{esfj}	&927	   &2,396,723\\
& \texttt{isfj}	&2,793 &8,970,103\\
& \texttt{istj}	&851 &2,373,061\\
\cline{2-4}
& \textbf{Subtotal} & \textbf{5,118} & \textbf{15,177,038}\\
\hline

\multirow{5}{*}{Explorers} 
& \texttt{isfp}	&6,345	&13,623,386\\
& \texttt{istp}	&1,258	&2,886,041\\
& \texttt{estp}	&1,032	&4,132,972\\
& \texttt{esfp}	&972  &3,831,533\\
\cline{2-4}
& \textbf{Subtotal} & \textbf{9,607} & \textbf{24,473,952}\\

\hline
& \textbf{Total} & \textbf{56,130} & \textbf{152,248,415}\\
\hline
\end{tabular}
\caption{The Twitter dataset. MBTI personality types have four dimensions. These are -  Extraversion (e) vs introversion (i) – where you get your energy from, Sensing (s) v intuition (n) – what kind of information you prefer to gather, Thinking (t) v feeling (f) – how you make decisions, and Judging (j) v perceiving (p) – how you deal with the world around you.\\
\# denotes `number of' in this table.
}
\label{tab:stats}
\end{table}

\subsection{Preprocessing}
We take the following preprocessing steps. 
\begin{enumerate}
    \item We detect the language of the tweets using \textbf{fasttext} and filter out all non-English tweets.
    \item  We keep only those users for whom at least 100 of the filtered tweets are present. This is necessary to obtain statistically meaningful results from the analysis we perform later. 

    \item We retain only unambiguous users, i.e., users having only one personality type. If a user tweets multiple `16personalities' links having different types, we remove them from our dataset to maintain consistency.

    \item We separate the hashtags, emojis, mentions, and URLs from the tweet text and analyze them individually. URLs may come from different media houses having different biases and attracting people of specific personalities; same may happen with hashtags, mentions and emojis\cite{souvic_media,souvic_ff}. Finally, we use TweetBERT's \textbf{tweet normalizer} to normalize the tweet text.
\end{enumerate}




\section{ Dataset Analysis}

MBTI personality types can be broadly mapped to four classes\cite{shestakevych2021modeling}. These are -  Extraversion (e) vs introversion (i) – where you get your energy from, Sensing (s) v intuition (n) – what kind of information you prefer to gather, Thinking (t) v feeling (f) – how you make decisions, and Judging (j) v perceiving (p) – how you deal with the world around you.
 
\begin{itemize}
\item \textbf{Analysts}: Intuitive (N) and Thinking (T).
\item \textbf{Diplomats}: Intuitive (N) and Feeling (F).
\item \textbf{Sentinels}: Observant (S) and Judging (J).
\item \textbf{Explorers}: Observant (S) and Prospecting (P).
\end{itemize}

The number of users and their tweets collected for each personality type is enumerated in Table \ref{tab:stats}. Further, we have done several quantitative and qualitative analyses of our large dataset using this mapping. These are summarized below.
 
\subsection{Readability metrics}
Readability is the ease with which a reader can understand a written text. In natural language, the readability of text depends on its content (the complexity of its vocabulary and syntax) and its presentation (such as typographic aspects that affect legibility, like font size, line height, character spacing, and line length) \cite{vstajner2012can}. We have calculated eight types of readability metrics for each user. These are: \textit{Flesch Readability} (\cite{flesch1948new}), \textit{Flesch-Kincaid Grade Level} (\cite{flesch2007flesch}), \textit{Dale Chall Readability} (\cite{williams1972table}), \textit{Automated Readability Index} (ARI \cite{senter1967automated}), \textit{Coleman Liau Index} (\cite{coleman1975computer}), \textit{Gunning Fog} (\cite{gunning1969fog}), \textit{Linsear-write} (\cite{eltorai2015readability}), \textit{SPACHE} (\cite{spache1973reading}). 

The average values of each personality class's readability metrics are shown in Table \ref{tab:read}. We can infer from the majority of readability metrics that tweets of \textbf{analyst} personality class are the hardest to read as compared to the other three. Similarly, \textbf{explorer} personality class tweets are the easiest to read among the three.

\begin{table}[htb]
  \resizebox{\linewidth}{!}{\begin{tabular}{l*{8}{c}}
    \toprule
    \textbf{Personality Class}
    & \textbf{flesch}&	\textbf{flesch-kincaid}	&\textbf{coleman-liau}	&\textbf{dale-chall}	&\textbf{gunning-fog}	&\textbf{ari}	&\textbf{linsear-write}&	\textbf{spache} \\ [0.5ex]
    \midrule
    \textbf{Analyst} &  \textbf{71.83}	&\textbf{5.29}	&\textbf{11.96}&	\textbf{13.49}	&\textbf{8.41}	&\textbf{8.17}&	\textbf{4.54}	&\textbf{7.32}\\
    \textbf{Diplomat} &  75.25	&4.77	&13.72	&14.14	&8.24	&9.63 &4.40	&7.57\\
    \textbf{Sentinel} &  75.09	&\textbf{4.71}	&13.90	&14.37	&\textbf{8.09}	&9.74	&\textbf{4.20}	&7.71\\
    \textbf{Explorer} &  \textbf{75.48}	&4.71	&\textbf{14.29}	&\textbf{14.94}	&8.31	&\textbf{10.13}	&4.35	&\textbf{7.94}\\

    \bottomrule

\end{tabular}}
\caption{Average readability scores - minimum and maximum values are highlighted in bold.}
\label{tab:read}
\end{table}

\subsection{Empath features}
 Empath features draw connotations between words
and phrases by learning a neural embedding from more
than 1.8 billion words of modern fiction, as proposed by \citet{fast2016empath}. Given a small set of
seed words that characterize a category, Empath uses its neural embedding to discover new related terms, then validates
the category with a crowd-powered filter. Empath also analyzes text across 194 built-in, pre-validated categories that the authors generated from common topics in their web dataset, like
neglect, government, and social media. We compute empath feature vectors for all users using the pre-trained model. The top distinct empath features for each personality class are presented in Table \ref{tab:empath}. We can see that most of them align with common intuition. For instance, while analysts correspond to words like `programming', `tool', and `optimism', explorers correspond to words like `dance', `music', and `appearance'. 

\begin{table}[ht]
\begin{tabular}{|c|c|c|c|}
\hline
\textbf{Analysts} & \textbf{Diplomats} &\textbf{Sentinels} & \textbf{Explorers}\\
\hline
optimism&	negotiate &	exercise&	dance\\
tool&	law&	messaging	&feminine\\
programming	& office &	white collar job&	music\\
government&	politics	&office	&cheerfulness\\
negotiate&	contentment	&blue collar job	&attractive\\
philosophy&	torment	&occupation	&appearance\\
law&	nervousness	&vehicle	&affection\\
\hline
\end{tabular}
\caption{Most distinct empath features in each of the personality classes.}
\label{tab:empath}
\end{table}



\subsection{Most distinct professions}
We find the most distinct professions for each personality class using the profile's description or biography. We first parse the descriptions into tokens and then calculate the probability of each personality class given that token, i.e., $Probability(Class | Token)$. We then take the words having high probability scores as the representative professions for a personality class. The results in Table~\ref{tab:accents} are quite intriguing and align with natural intuition. Analysts have distinctive professions like fullstack engineer and scientist, Diplomats have campaigner and theorist and Sentinels have surgeon and dentists. We do not find any such alignments for Explorers as they may have a diverse range of professions and they do not significantly identify themselves with one profession but they do explore some newer professions driven by social media and web 3.0.  


\subsection{Metadata statistics}
\begin{figure}[h]
     \centering
     \begin{subfigure}[b]
        \centering
        \includegraphics[width=0.3\textwidth]{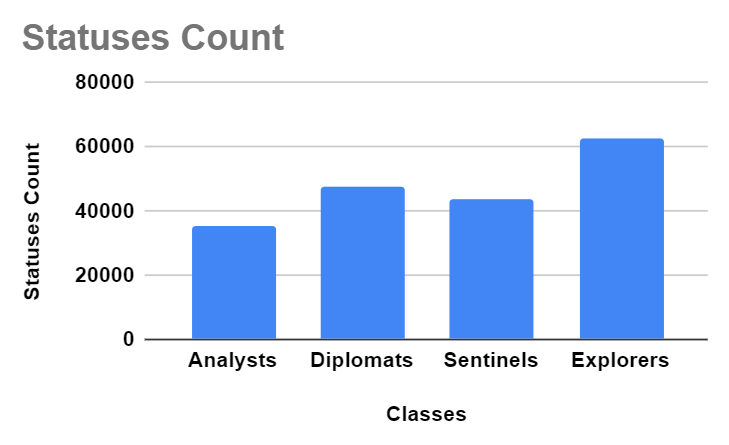}
        \caption{Average statuses count.}
        \label{fig:statuses}
    \end{subfigure}

\begin{subfigure}[b]
        \centering
        \includegraphics[width=0.3\textwidth]{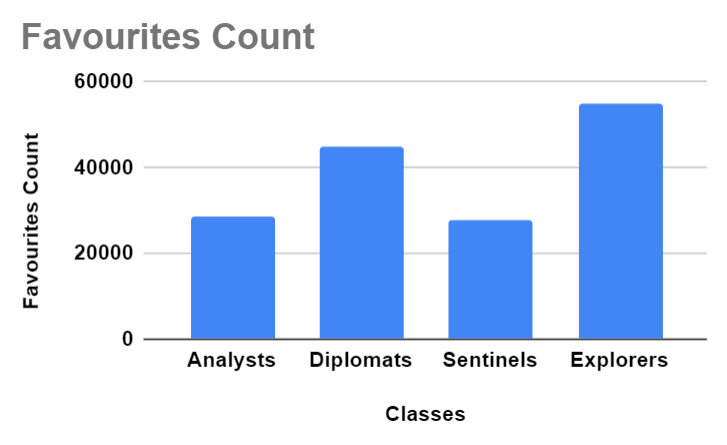}
        \caption{Average favorites  count.}
        \label{fig:favor}
    \end{subfigure}

\begin{subfigure}[b]
        \centering
        \includegraphics[width=0.3\textwidth]{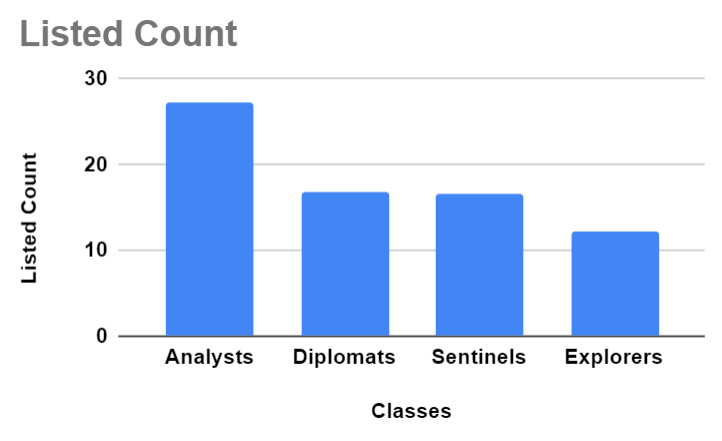}
        \caption{Average listed count.}
        \label{fig:list}
    \end{subfigure}
\caption{Metadata statistics of the dataset for each personality class.}\label{fig:metadata}
\end{figure}

From the metadata statistics of the user profiles as shown in Figure~\ref{fig:metadata}, we observe that the explorers update their statuses the most and analysts update the least.
The favorites count is the highest for the explorers followed by the diplomats.
The listed Count shows how many people have added the user to a list. 
We see that \textbf{analysts} are the most listed personality type, followed by diplomats, sentinels, and explorers. This is probably because analysts are good at rational thinking and can explain complex information in a way that is easy to understand. These traits are beneficial for using Twitter, as the platform requires concise and effective communication. 

\begin{table}
\centering
\begin{tabular}{|c|c|c|}
\hline
\textbf{Analysts} & \textbf{Diplomats} &\textbf{Sentinels}\\
\hline
fullstack	&socialist	&epidemiologist \\
systems&	campaigner&	accounting \\
trader&	novelist	&surgeon \\ 
astrology&	theorist&	dentist \\ 
scientist & scientist &oncology \\
\hline
\end{tabular}
\caption{Most distinct professions per personality class.}
\label{tab:accents}
\end{table}

\begin{figure*}[ht]
    \centering
    \includegraphics[width=0.8\textwidth]{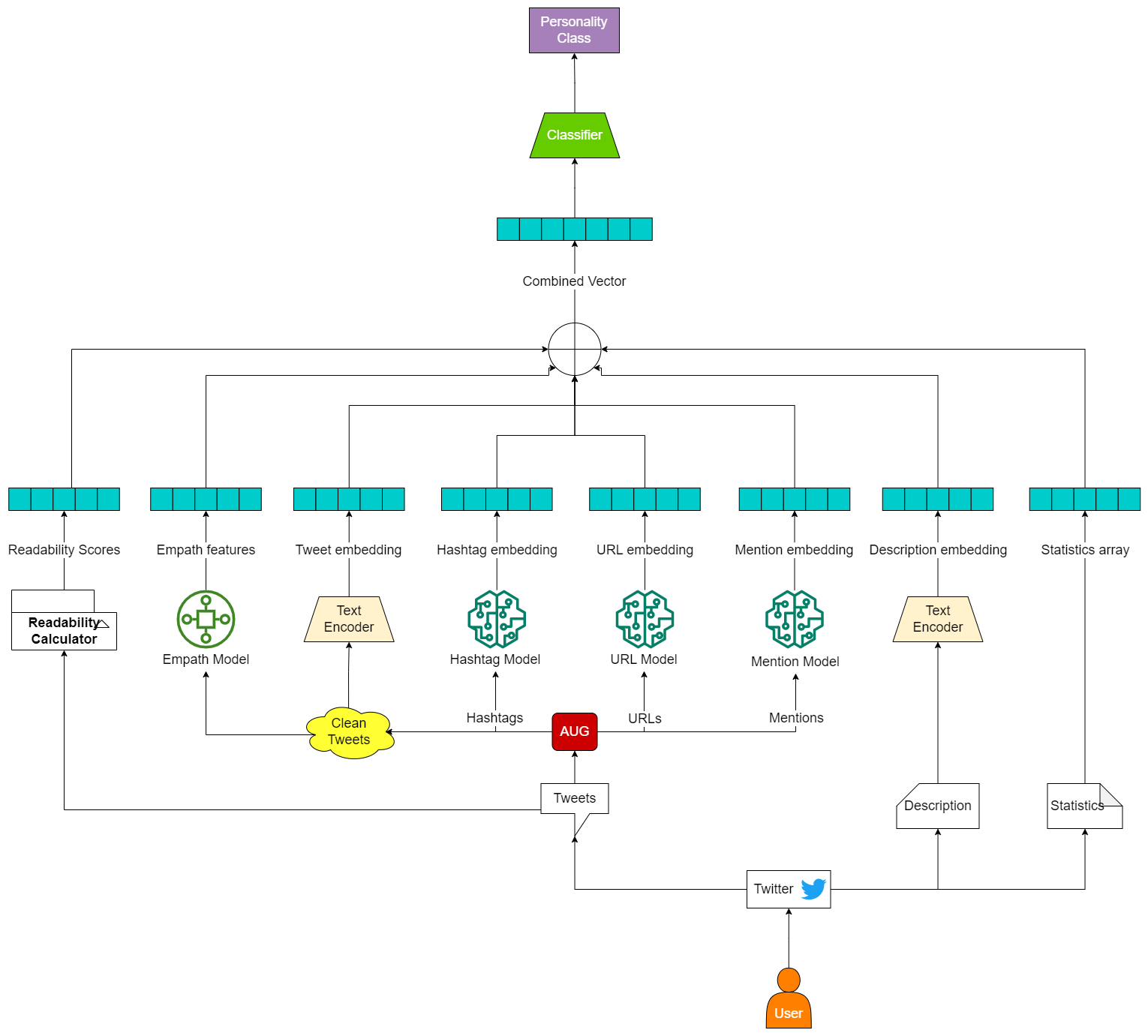}
    \caption{The overall architecture of our model. The AUG module separates the hashtags, URLs and mentions, and cleans the rest of the tweets. }
    \label{fig:model_arch}
\end{figure*}

\section{Methodology}

Our task is to classify Twitter users into four personality types namely discussed in the previous sections. The input features of the users available to us are - 1) the latest 3200 tweets, 2) the bio (description), and 3) profile statistics (follower count, media count, listed count, etc.). We compute eight different readability metrics from the tweets defined in the previous section. While preprocessing, we clean the tweets and store the hashtags, URLs, and mentions separately. We then compute the empath features using a pre-trained model from the clean tweets. We also embed hashtags, URLs, and mentions, as they may contain valuable information about a user's personality.

\subsection{URL, hashtag, and mention embeddings}

It has been seen that hashtags contain very indicative and valuable information about the user's personality. To capture this information, we calculate embeddings for the hashtags present in a user's tweets. We first concatenate the hashtags and vectorize them using \texttt{tf-idf}. We ignore all tokens which appear in less than 2\% of the tweets. We then pass the vectors to a neural network containing three dense layers and try to predict the personality class. After proper training, we use the output of the second last layer of the neural network as the user embedding for the hashtags. We follow a similar procedure for computing the URL and mention embeddings for each user.


\subsection{Approach}
To classify each user, we use a similar methodology for all the baselines. After preprocessing, we encode the tweets and descriptions using an encoder (\texttt{fasttext}, \texttt{bert}, \texttt{tweetbert}, and \texttt{roberta}). Then we vectorize the empath features, readability scores, and Twitter profile statistics (counts), and concatenate all these vectors. Finally, we concatenate the URL, hashtag, and mention embeddings with them. We use nine different configurations for classification depending upon the features chosen for input which are as follows. Our objective here was to understand the impact of each feature.
\begin{enumerate}
    \item All the features.
    \item Only the tweets.
    \item Without the URL embeddings.
    \item Without the hashtag embeddings.
    \item Without the mention embeddings.
    \item Without the URL, hashtag, and mention embeddings.
    \item Without the readability scores.
    \item  Without the empath features.
    \item Without the profile statistics.
\end{enumerate}

 For the classification of tasks, we have used various machine learning models which are described in the next section. We create a class-balanced subset of the whole data, containing approximately 4000 data points per class, sampled randomly for model training purposes. For the testing of models, we fix a random sample of approximately 1000 data points from each class from the remaining data. This ensures unbiased training and proper evaluation of the models. Our model architecture is shown in Figure~\ref{fig:model_arch}.

\begin{table*}[htb]
  \resizebox{\linewidth}{!}{\begin{tabular}{l*{16}{|c}|}
    \toprule
    & \multicolumn{4}{c|}{\texttt{fasttext} \textbf{embeddings}} 
    & \multicolumn{4}{c|}{\texttt{bert} \textbf{embeddings}}
    & \multicolumn{4}{c|}{\texttt{tweetbert} \textbf{embeddings}}
    & \multicolumn{4}{c|}{\texttt{roberta} \textbf{embeddings}}\\ Configuration
    & \multicolumn{2}{c|}{\textbf{RFC}} & \multicolumn{2}{c|}{\textbf{XGB}}
    & \multicolumn{2}{c|}{\textbf{RFC}} & \multicolumn{2}{c|}{\textbf{XGB}} & \multicolumn{2}{c|}{\textbf{RFC}} & \multicolumn{2}{c|}{\textbf{XGB}} & \multicolumn{2}{c|}{\textbf{RFC}} & \multicolumn{2}{c|}{\textbf{XGB}}
     \\ [0.5ex]
    & F1 & Acc & F1 & Acc & F1 & Acc & F1 & Acc & F1 & Acc & F1 & Acc &F1 & Acc & F1 & Acc  \\
    \midrule
    All features &   \textbf{45.71}	&  46.03 &  44.40&  44.66& 42.86& 43.86& \textbf{43.14} &  43.39&  \textbf{42.21}& 43.36& 42.26 &  42.68 & \textbf{43.06} & 44.06 & \textbf{42.02} & 42.41\\
    
   W/o URLs	& 45.13	& 46.38 &  44.18&  44.29&  42.28& 43.28&  42.94& 43.18 & 41.96& 43.06 & 42.15 &  42.34 & 42.05 & 43.18 & 41.67 & 42.03\\
   W/o (\#)&	 43.45&	 43.76&  43.40&  43.66& 42.15& 43.21&  42.59& 43.38 &  41.01& 42.11 & 40.96 &  41.98 & 42.36 & 43.51 &41.67 & 42.03\\
   W/o (@)&	 44.13	& 45.11&  44.20 &  43.56&  42.32& 43.43&  42.15& 42.68 &  41.70& 42.78 & 41.16 &  41.38 & 42.17 & 43.26& 41.67 & 42.03\\
   W/o URLs,@,\# &   43.92	&  44.22 &  43.18&  42.56& 41.26& 41.34& 41.15 &  41.29&  41.21& 42.16& 40.86 &  41.18 & 42.65 & 43.63 & 41.67 & 42.03\\
Only tweets	&  45.24	& 46.33&  44.21& 44.56&   40.55& 41.58&  41.12 & 41.36&  40.34& 41.43& 39.31 &  39.66 & 40.85 & 41.86 & 40.05 & 40.36\\

W/o readability& 45.55	& 46.73&  45.36 &  45.64&  \textbf{42.95}& 43.88& 42.21 & 42.51&  41.56& 42.63&  \textbf{42.82} &  43.13 & 41.81 & 42.93 & 41.84 & 42.18\\
W/o counts&	 45.27&	 46.63&  \textbf{45.55}&  45.84&  42.04& 43.18&  42.87 & 43.19&  41.60& 42.66&  42.65 &  42.96 & 42.83 & 43.81 & 41.67 & 42.03\\
W/o empath&	44.76&	46.11& 43.61& 44.24& 40.53&41.63&41.83 &42.06& 40.24&41.26& 40.59 & 41.03 & 41.20 & 42.18 & 40.20 & 40.53\\
    \bottomrule

\end{tabular}}
\caption{Overall results for the four-class classification task. The best F1 scores obtained are highlighted in bold.}\label{tab:5}
\end{table*}

\subsection{Baselines}
We use four different encoders to encode the tweets and descriptions into embeddings - \texttt{fasttext}, \texttt{bert}, \texttt{tweetbert}, and \texttt{roberta} For classification, we use two classical models - Random Forest and XGBoost. We also employ other machine learning as well as deep learning variants; however, the results being poorer we refrain from reporting them in the paper.

We concatenate the output of the encoder in each case with the empath (194 dim), readability (8 dim), metadata counts (6 dim), and the mention, hashtag, URL embeddings (64 dim each) to get the final feature vector which is used for classification using one of the algorithms mentioned above.

We use default hyperparameters of \textit{sklearn} and \textit{fasttext} libraries in every case for finetuning here as they presented the best results.

\subsubsection{Embeddings}
\begin{enumerate}
\item \texttt{Fasttext}: FastText\cite{joulin2016fasttext} is an open-source, free library from Facebook AI Research for learning word embeddings and word classifications. We use pre-trained fastText embeddings to convert the tweets of each user and their descriptions into 700-dimensional vectors.
\item \texttt{Bert}: Bidirectional Encoder Representations from Transformers is a transformer-based machine learning technique for natural language processing pre-training developed by Google\cite{devlin2018bert}. To encode the tweets, we first tokenize the last 64 tokens using the BERT tokenizer, then pass it into the \texttt{bert-base} model to get a single tweet embedding. Likewise, we do the same for all 3200 tweets, and take an average to get the tweet embedding for a single user.
\item \texttt{Tweetbert}: TweetBERT is a BERT model that has been trained on Twitter datasets, and shows significantly better performance on text mining tasks on Twitter datasets (proposed in \citet{qudar2020tweetbert}). We follow a similar strategy as in the case of BERT to obtain the tweet embeddings.
\item \texttt{Roberta}: A Robustly Optimized BERT Pretraining Approach (RoBERTa) was proposed by \citet{liu2019roberta}. It builds on BERT and modifies key hyperparameters, removing the next-sentence pretraining objective and training with much larger mini-batches and learning rates. We follow a similar strategy as in the case of BERT to obtain the tweet embeddings.

\end{enumerate}
\subsubsection{Classifiers}
\begin{enumerate}
\item \textbf{Random forest classifier} (RFC): The random forest classifier is commonly used to reduce variance within a noisy dataset. It significantly raises the stability of models by improving accuracy and reducing variance, which eliminates the challenge of overfitting. An improved version of the classifier was proposed by \citet{xu2012improved} which we use for our experiments.
\item \textbf{Extreme gradient boosting} (XGB): The gradient boosting classifier model helps in reducing variance and bias in a machine learning ensemble. An efficient and scalable
implementation of gradient boosting framework is developed by \citet{chen2015xgboost}, which is called XGBoost. We use this model for all our experiments.
\end{enumerate}

\section{Results}

The main results from the classification are presented in Table~\ref{tab:5}. 
From the table, we see that RFC performs better than XGB in most of cases. The most important features are the \textbf{tweets} and \textbf{hashtag embeddings}. The effect of the other features is minimal. As for the embedding learning algorithms, we observe that all of them perform similarly, with a small edge going to the \texttt{fasttext} encoder. The best F1 score is reached by using one of (i) \texttt{fasttext} embeddings with RFC and all features or (ii) \texttt{fasttext} embeddings with XGB and all but the profile statistics features.

 Our results further show that employing the URL, hashtag, and mention embeddings along with all other features (readability, counts, etc) gives an overall boost of $\sim1-2\%$ in terms of classification F1 score, while the use of profile statistics (followers, listed count, etc.) also gives an overall boost of $\sim0.5\%$. The use of the empath features gives an overall boost of $\sim0.9-1\%$ in the F1 score and accuracy. The readability scores showed some improvement in accuracy and F1 score for some pairs of encodings and models, although it had the least contribution compared to the other features.

\section{Error analysis}
Detecting MBTI personality types can introduce several possibilities of errors. Some are as follows.
\begin{itemize}
    \item \textbf{Lack of control over the sample population}: People who use Twitter are not necessarily representative of the general population. There may be biases in terms of age, gender, ethnicity, and socio-economic status, which can impact the accuracy of the MBTI personality type identification. 
    \item \textbf{Variability in expressing personality traits}: Individuals can express their personalities in different ways depending on the situation or context. For example, individuals who are typically introverted may appear to be extroverted in certain social settings. 
    \item \textbf{Difficulty in measuring some personality traits}: The MBTI measures personality traits that are not necessarily easily observable, such as intuition or sensing. Moreover, these traits are not always consistently displayed in tweets.
\end{itemize}

Using tweets to detect MBTI personality types is an interesting and innovative approach but the above limitations can introduce inaccuracies and errors in the predictions. To illustrate this we present some examples of predictions done by our best model in Table \ref{tab:error}. Our model does well when the tweets reflect a single trait in their behavior clearly. However, it commits errors when the information available is confusing. For instance, note that since User 5 is a YouTuber (in addition to being an athlete) so our model predicts the person to be an explorer (which is, in fact, partially correct). The case of User 6 is more common -- many actors in the later stage of their careers enter into active politics (e.g., Hema Malini\footnote{\url{https://en.wikipedia.org/wiki/Hema_Malini}}, J. Jayalalithaa\footnote{\url{https://en.wikipedia.org/wiki/J._Jayalalithaa}}, Clay Aiken\footnote{\url{https://en.wikipedia.org/wiki/Clay_Aiken}}, Alessandra Mussolini\footnote{\url{https://en.wikipedia.org/wiki/Alessandra_Mussolini}}, Maria Kozhevnikova\footnote{\url{https://en.wikipedia.org/wiki/Maria_Kozhevnikova}}, Jimmy Edwards\footnote{\url{https://en.wikipedia.org/wiki/Jimmy_Edwards}}, etc.). Our model finds it hard to classify such cases probably because while they self-report themselves as diplomats, they still tweet a lot about the acting world. The error in the case of User 7 arises because the person writes very complex tweets which is an unusual trait for explorers and a usual trait for analysts. The last case (User 8) is also confusing since the person tweets, tags, and mentions political entities. Thus, in summary, while a person's personality class is usually thought to be fixed there might be cases where it can branch out due to multiple interests pursued or due to followership of an ideology or a school of belief, or due to a change in the profession over time. Therefore the model predictions should always be used with appropriate caution.

\begin{table*}
\centering
\begin{tabular}{|c|c|c|c|}
\hline
\textbf{User} & \textbf{MBTI Class} &\textbf{Predicted Class} & \textbf{Reason}\\
\hline
User 1	& Analyst	&Analyst & The user is a part of cryptography learning community \\
User 2 & Diplomat &	Diplomat & The user has 'Idealist' in his profile description (bio) \\
User 3 & Explorer &	Explorer & The user is an Actress \\ 
User 4 & Sentinel &Sentinel & The user is a Dentist\\ 
User 5 & Sentinel &Explorer & The user is an athlete as well as a YouTuber \\
User 6 & Diplomat &Explorer & The user is an actor turned politician \\
User 7 & Explorer & Analyst & The user writes very complex posts \\
User 8 & Analyst & Diplomat & The user is politically active and mentions a lot of politicians \\
\hline
\end{tabular}
\caption{Some examples of correct and incorrect cases, and the reasoning behind the prediction.}
\label{tab:error}
\end{table*}

\section{Conclusion}
In this work, we released the largest automatically curated Twitter dataset for personality detection for MBTI personality types. Then we classified Twitter users into personality types - analysts, diplomats, sentinels, and explorers using the latest 3200 tweets and profile information. We derived new features from the tweets to capture user personality, as well as computed embeddings from the URLs, hashtags, and mentions. We used various encoders (FastText, BERT, TweetBERT, and RoBERTa) to convert the tweets into embedding vectors followed by traditional machine learning models for classification.

\section{Limitations}

Human language is highly dynamic. Most of the metadata present in the tweets such as hashtags and mentions touches upon topics and their difficulty which may not be well represented by existing machine learning models. In addition, even though we incorporate readability metrics, it may still not be enough to capture an individual’s attitude and behavior accurately. Also, due to the length constraint of tweets, deeper context cannot be extracted easily. Finally, our model may not be able to account for changes in users' personalities over time.

\section{Future Works}
 The task of classifying Twitter users according to their personality type is an interesting research area, with many potential applications. We believe that there is room for improvement in our existing method in terms of accuracy and runtime. While text-based data such as tweets can provide valuable insights into a user's personality, incorporating audio and video data from social media platforms such as YouTube and TikTok could provide additional information. However, analyzing such data can be challenging due to its unstructured nature, making this a potentially challenging future work. Further, social media users' personalities can evolve and change over time, making it difficult to classify them accurately based on a single snapshot of their behavior. Developing a model that can capture temporal dynamics and classify users based on their personality over time could be another future direction. Further, since the challenge with large models is interpretability, we would also like to investigate this avenue by digging deeper into the relationships among the input features. In addition, we would also like to explore the potential of multi-dimensional classification to provide more granular information about the personality type of a Twitter user. The current models could also be extended to other social media platforms such as YouTube and Instagram, as the personalities of users on these platforms could influence the type of content they generate and hence could indicate their MBTI type.

\bibliographystyle{unsrtnat}
\bibliography{sample-base}
\end{document}